%% file: StyleLess_ARXIV_2021.tex
\documentclass[letterpaper, 10 pt, conference]{ieeeconf} 
\usepackage{graphics}
\usepackage[utf8]{inputenc}
\usepackage[T1]{fontenc}
\usepackage{hyperref}
\usepackage{url}
\usepackage{booktabs}
\usepackage{amsfonts}
\usepackage{nicefrac}
\usepackage{microtype}
\usepackage{graphicx}
\usepackage{hhline}
\usepackage{bmpsize}
\usepackage{subfigure}
\usepackage{amssymb}
\usepackage{amsmath}
\usepackage{dsfont}
\usepackage{multirow}
\usepackage{booktabs} 
\usepackage{stmaryrd} 
\usepackage{xcolor}
\usepackage{flushend}
\usepackage{textcomp}
\usepackage{units}
\usepackage{caption}
\usepackage{algorithm}
\usepackage[noend]{algorithmic}
\usepackage{mdframed}
\usepackage[normalem]{ulem}
\usepackage{wrapfig}
\usepackage{adjustbox}
\usepackage{xspace}
\usepackage{siunitx}

\input{abbrev.tex}

\title{\LARGE \bf
StyleLess layer: Improving robustness for real-world driving
}

\author{Julien Rebut, Andrei Bursuc, and Patrick Pérez\\
valeo.ai
}

\begin{document}

\maketitle
\thispagestyle{empty}
\pagestyle{empty}

\begin{abstract}
Deep Neural Networks (\dnns) are a critical component for self-driving vehicles. They achieve impressive performance by reaping information from high amounts of labeled data. Yet, the full complexity of the real world cannot be encapsulated in the training data, no matter how big the dataset, and \dnns~ can hardly generalize to unseen conditions. Robustness to various image corruptions, caused by changing weather conditions or sensor degradation and aging, is crucial for safety when such vehicles are deployed in the real world. We address this problem through a novel type of layer, dubbed \emph{\method}, which enables \dnns~to learn robust and informative features that can cope with varying external conditions. 
We propose multiple variations of this layer that can be integrated in most of the architectures and trained jointly with the main task. 
We validate our contribution on typical autonomous-driving tasks (detection, semantic segmentation), showing that in most cases, this approach improves predictive performance on unseen conditions (fog, rain), while preserving performance on seen conditions and objects.
\end{abstract}

\input{1-introduction.tex}
\input{2-related-work.tex}
\input{3-approach.tex}
\input{4-experiments.tex}
\input{5-conclusion.tex}

\addtolength{\textheight}{-2.2cm}
{\small
\bibliographystyle{ieee}
\bibliography{StyleLess_ARXIV_2021}
}
\end{document}

%% file: abbrev.tex
\newcommand*\rot{\rotatebox{90}}

\newcommand{\method}{StyleLess}
\newcommand{\dnn}{DNN}
\newcommand{\dnns}{DNNs}

\newcommand{\mean}{\operatorname{mean}}

\newcommand{\diag}{\operatorname{diag}}

\newcommand{\real}{\mathbb{R}}

\newcommand{\new}[1]{#1'}

\newcommand{\nH}{\new{H}}
\newcommand{\nW}{\new{W}}

\newcommand{\nh}{\new{h}}
\newcommand{\nw}{\new{w}}

\newcommand{\norm}[1]{\left\|{#1}\right\|}
\newcommand{\frob}[1]{\norm{#1}_F}

\newcommand{\vx}{\mathbf{x}}

\newcommand{\vtheta}{\boldsymbol{\theta}}
\newcommand{\vpsi}{\boldsymbol{\psi}}

\newcommand{\vzero}{\mathbf{0}}
\newcommand{\vepsilon}{\boldsymbol{\epsilon}}

\newcommand{\vsigma}{\boldsymbol{\sigma}}

\newcommand{\parag}[1]{\smallskip\noindent\textbf{#1}~~}

\makeatletter
\DeclareRobustCommand\onedot{\futurelet\@let@token\@onedot}
\def\@onedot{\ifx\@let@token.\else.\null\fi\xspace}

\def\ie{\emph{i.e}\onedot} 
 
\def\etc{\emph{etc}\onedot}

\makeatother

%% file: 1-introduction.tex
\section{INTRODUCTION}
\label{Introduction}
The tremendous progress in computer vision and machine learning in recent years translated into new applications and functionalities for autonomous systems. Cameras are key and cost-effective sensors enabling \dnn-based visual perception for driving assistance or for autonomous driving. An excellent trade-off between affordability and perception performance enables cameras to be deployed at large scale in most of the new vehicles across various price ranges and levels of intervention and autonomy. 

During the life-cycle of the vehicle, about 10-15 years, the camera is expected to encounter a plethora of diverse weather and lighting conditions, traffic scenarios, road users in an ever changing world
(e.g., pedestrians with masks, scooters on the road, new car models, headlamps changing from halogen to LED, \textit{etc}.), while undergoing  gradual hardware degradation due to aging with a non-negligible impact in perception performance. 
To some extent, \dnns~can address this by learning from large data repositories recorded with numerous vehicles and by integrating different training heuristics for improving robustness, e.g., data augmentation~\cite{hendrycks2019augmix} and adversarial training~\cite{hendrycks2019using}. However the full complexity of the world and its evolution along with various types of sensor degradation cannot be encapsulated in the training data at a given time snapshot, no matter how big the dataset. This challenge is currently addressed with regular \emph{over-the-air} updates, usually on premium vehicles, where better performing models trained on recent data and on additional scenarios are downloaded. This strategy could also be conceivable for some mid to lower-end car models, however updates might be less frequent due to the large diversity of fleets of different ages, while the weaker hardware will likely be constraining after a number of years, further reducing the extent of the updates. In this work we aim to circumvent this problem by training \emph{robust} \dnns~to cope with unseen external conditions and potential visual perturbations caused by camera aging or deterioration.

\begin{figure}[t!]
    \renewcommand{\captionfont}{\small}
    \centering
    \includegraphics[width=8cm]{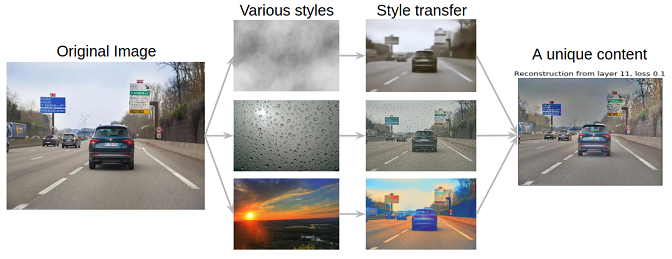}
    \caption{When neural style transfer~\cite{gatys2016image} is performed over the image on the left using different style images (fog, rain, sunset), this results in images with different visual appearance, \ie the \emph{style reconstruction loss} is high between them. However the \emph{content reconstruction loss} is close to zero, meaning that we could 
    learn a \emph{style-less} representation invariant to visual perturbations. Learning such representation is the main objective of this work.
    }
    \label{StyleTransfer}
\vspace{-10pt}
\end{figure}

We first define \emph{robustness} for the scope of this work: \emph{a system is robust if its predictions remain unchanged when the input signal is subject to perturbations}. Multiple works have studied different flavors of \emph{robustness} in the past years. Robustness to adversarial attacks~\cite{szegedy2013intriguing,papernot2016practical,eykholt2017robust} has  been extensively studied mostly from a security viewpoint with several directions for improving or attacking \dnns~\cite{goodfellow2014explaining, carmon2019unlabeled}.
The scope of robustness has been subsequently enlarged with other types of failures of \dnns: spurious correlations between objects and context~\cite{Elephant, shetty2019not},bias towards textures~\cite{geirhos2018imagenettrained}, objects with misleading visual aspect~\cite{hendrycks2019natural}, unseen hazardous situations or objects~\cite{Zendel_2018_ECCV, hendrycks2019benchmark}, errors identified via systematic testing~\cite{pei2017deepxplore, tian2017deeptest} or image corruptions~\cite{hendrycks2019benchmarking, pezzementi2018putting}. In this work, we address robustness through the lens of an additive noise over feature maps. 

In contrast to prior works~\cite{tian2017deeptest, pezzementi2018putting, geirhos2018imagenettrained} where the noise signal is leveraged for improved data augmentation of the input, we propose a method that acts in the feature space. We argue that even though several realistic perturbations, e.g., fog~\cite{dai2019curriculum} or rain~\cite{RainyCityscapes, tremblay2020rain}, of the training data can be performed, designing a rich set of perturbations covering all sorts of environment variations and sensor specific deterioration, e.g., camera, radar, lidar, \etc.,  is a daunting task. We build on findings from neural artistic style transfer~\cite{gatys2016image}, where ``style'' (colors, textures) and ``content'' (geometric layout) information can be manipulated via \dnn~features. We propose a new \method~layer that reduces style information from features and enables the network to focus instead on the content, which is preserved across varying environment conditions.

\noindent \textbf{Contributions.} To summarize, the contributions of our work are: (1) We propose a pragmatic definition of robustness and connect it with style transfer where both style and content information can be interpreted. (2) We design a new layer, \method, that enables the model to ignore noisy style information and learn from the content instead; \method~does not depend on the input sensor and can be extended to other sensors coupled with \dnns. (3) We evaluate our method on autonomous-driving tasks, object detection and semantic segmentation, in challenging test conditions with adverse weather and corrupted images showing superior results over methods trained with sophisticated data augmentation.

%% file: 2-related-work.tex
\section{Related Work}
The field of automatic visual perception has seen significant progress with the recent wave of \dnn~approaches and architectures~\cite{ szegedy2014going, he2016deep}.
The impressive performance of \dnns~on popular academic benchmarks encouraged their transfer towards practical applications, e.g., traffic surveillance, driving assistance, autonomous systems,  \etc. In contrast with typical benchmarks where the test set is fixed and usually follows a similar distribution with the train set, in real-world applications the test set is practically changing continuously and robustness to such changes and unexpected situations is essential for the reliability of such systems. Robustness has thus gathered interest from different research communities ranging from machine learning~\cite{raghunathan2020understanding, Gowal19} to automatic testing~\cite{pei2017deepxplore, tian2017deeptest}. 

Robustness against adversarial attacks~\cite{szegedy2013intriguing,papernot2016practical,eykholt2017robust} is one of the most studied aspects of \dnns~due to the spectacular failures exposed by such techniques~\cite{brown2017adversarial, cisse2017houdini}.
We focus on robustness related to perception of the visual world, rather than blind spots identified through targeted attacks.
The study of robustness has been 
later extended to identifying prediction errors due to inherent bias in the training set such as the influence of context and object co-occurrence in predictions~\cite{Elephant, shetty2019not, srivasta2020human}, e.g., an elephant pasted over an image of a living-room is no longer detectable. Another manner to evaluate robustness is by studying \dnn~predictions when dealing with unseen objects~\cite{hendrycks2019benchmark} or unseen hazardous situations~\cite{Zendel_2018_ECCV}. This line of works addresses robustness from an epistemic perspective~\cite{kendall2017uncertainties}, which can be partially mitigated with additional training data. Here, we address rather the robustness w.r.t. variations in the visual appearance of the environment and objects.

Recently, Geirhos et al.~\cite{geirhos2018imagenettrained} pointed out that ImageNet-pretrained networks have a strong bias towards textures, unlike humans who tend to focus on shapes. Such networks over-rely on textures, while being agnostic to the object representation as a whole. 
This bias can be mitigated by augmenting the train set with stylized versions~\cite{gatys2016image} of the train images, leading to increased robustness to noise.
Kamann et al.~\cite{kamann2020increasing} build upon this finding and propose a simple method for increasing shape bias, while foregoing the high computational overhead of the style-transfer augmentation. They propose a data augmentation technique for semantic segmentation that leverages segmentation masks to add class and instance specific noise, e.g., color alpha-blending, to objects in the scene. 
They show that for semantic segmentation, style-transfer augmentation turns out to be a less suitable strategy due to the structured-output nature of the task and a potential mismatch of artistic styles for autonomous driving datasets. Conversely, Michaelis et al.~\cite{michaelis2019benchmarking} show that stylized augmentation actually helps in their robust object detection benchmark, leading us to conjecture that the style-transfer strategy should be elaborated according to both the dataset, the task and the sensor.
Our \method~layer approach enables the network to reduce style bias during training and does not rely on a specific data augmentation strategy, that usually needs data expertise and tuning. \method~layer can be extended to other modalities that are processed with \dnns.

Some works emphasize the evaluation of the robustness through image perturbations and identification of corner cases that lead the system to fail~\cite{pei2017deepxplore, tian2017deeptest}, cases that are then re-injected in the training pipeline. In this direction, image datasets with varying levels of corruption have been proposed to evaluate robustness~\cite{hendrycks2019benchmarking}. However these perturbations are synthetic and do not always resemble to practical edge cases. New driving-specific cases have been studied recently focusing on different outdoor conditions, e.g., rain~\cite{ICRA19_porav}, fog~\cite{FoggyCityscapes}, nighttime~\cite{sakaridis2019lowlight}. They reveal that data augmentation may not always be the best strategy, being second to image restoration~\cite{ICRA19_porav}, curriculum model adaptation~\cite{dai2019curriculum, sakaridis2019lowlight} or unsupervised domain adaptation~\cite{advent}. Contrarily to these strategies that make use of adverse weather data during training, our approach enables learning robust representations without such data and specific assumptions.

%% file: 3-approach.tex
\begin{figure*}[t!]
\renewcommand{\captionfont}{\small}
\centering
\includegraphics[width=0.85\linewidth]{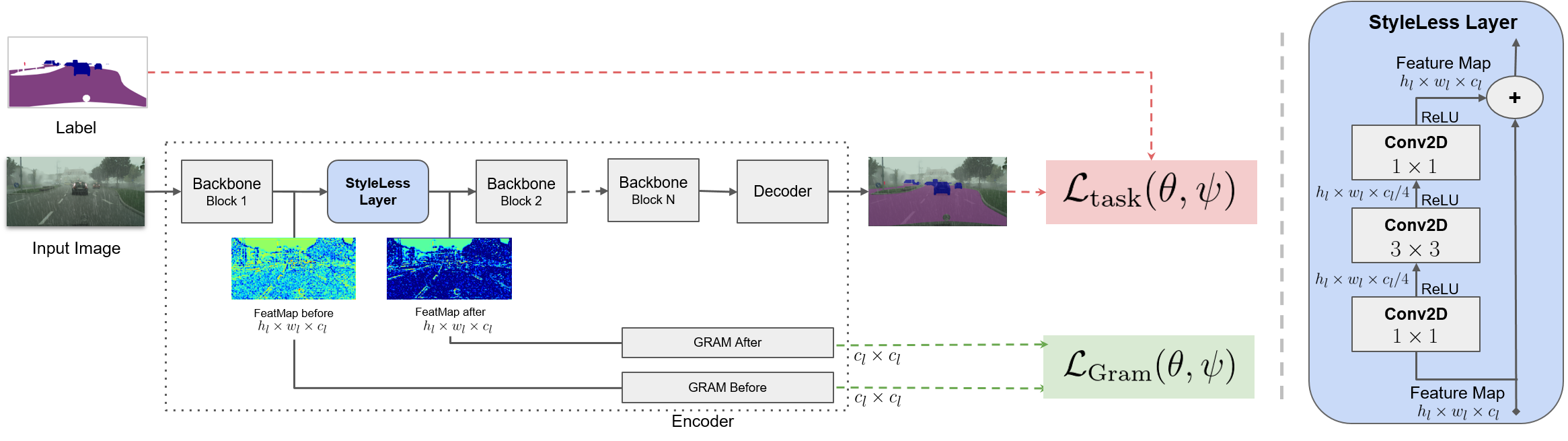}
  \caption{\textbf{Approach overview.} The figure shows our approach for training networks with robust representations. \emph{Left:} \method~layers are integrated at multiple intermediate points in the network and activated during the fine-tuning. The network is trained to minimize simultaneously the task loss $\mathcal{L}_{\text{task}}$ and the Gram loss $\mathcal{L}_{\text{Gram}}$, thus removing useless style information (see a typical example of the impact of a \method~layer over an input feature map) while maximizing prediction performance. \emph{Right:} Implementation of our \method~layer mimicking a residual block.
  } 
  \label{architecture}
\vspace{-10pt}
\end{figure*}

\section{Approach}

Our approach builds upon the observation that the world surrounding a vehicle is structured and the driving scene contains two main types of information: \emph{hard} information related to e.g., road, buildings, obstacles, and \emph{soft} information related to e.g., weather conditions, illumination, \etc. We can see the soft information as a ``filter'' that affects the perception of the world. In autonomous driving, we are interested in learning to grasp the hard information and rely to a smaller extent on soft information. In other words, we seek to learn robust features invariant to changes in external conditions. 
\subsection{Style and content information}
We can relate our observation on hard/soft information to the findings in the \emph{style transfer} task in computer vision~\cite{gatys2016image,johnson2016perceptual}, where an image's visual appearance, called ``style'' (essentially the colors and textures), is transformed to resemble another image, while preserving the original ``content'' (semantic and geometric layout). Gatys et al.~\cite{gatys2016image} showed the style and content information from an image can be decomposed and manipulated by processing DNN features. The content information is carried in the higher layers of the network, while the style can be recovered from feature correlations of multiple layers. In the following we provide background on neural style transfer~\cite{gatys2016image} and then  describe our approach that is built on these findings.

In style transfer, an input image $\vx \in \real^{H\times W \times 3}$ is given the style of another image $\vx_{\text{style}} \in \real^{\nH \times \nW \times 3}$ while preserving the original content in $\vx$. We denote the new stylized image by $\hat{\vx}$. The content and style information can be computed from the features of a pre-trained neural network $f_{\vtheta}(\cdot)$ 
with $L$ layers and parameters $\vtheta$. We denote by $F^{(l)}$, $\hat{F}^{(l)} \in \real^{(h_{l}w_{l}) \times c_{l}}$ and  $F_{\text{style}}^{(l)} \in \real^{(\nh_{\ell} \nw_{l}) \times c_{l}}$ the features at the $l^{\text{th}}$ layer of $f_{\vtheta}$ for the images $\vx$, $\hat{\vx}$ and $\vx_{\text{style}}$, respectively. 

The content information is carried in the hidden feature maps $F^{(l)}$ from top layers, preserving the spatial information, and thus is useful for scene understanding. On the other hand, the style captures texture and visual information regardless of the spatial layout of the scene. 
The style is encoded in the Gram matrices $G^{(l)}{=}F^{(l)\top} F^{(l)} / (h_{l} w_{l} c_{l}) $, computed at multiple layers and integrating correlations between different feature maps\footnote{The matrix $G^{(l)}$ is square of size $(c_{l}{\times}c_{l})$ where $c_{l}$ is the number of channels at feature map $l$.}. The style can be interpreted as a signal that modifies the appearance, but not the content itself. 

Neural style transfer~\cite{gatys2016image} amounts to minimizing w.r.t.  
image $\hat{\vx}$ the following function of the three images' features: 
\begin{align}
\mathcal{L}(\hat{\vx}) & = \mathcal{L}_{\text{content}}(\hat{\vx}, \vx) + \mathcal{L}_{\text{style}}(\hat{\vx}, \vx_{\text{style}}),\\
\mathcal{L}_{\text{content}}(\hat{\vx}, \vx) & = \sum_{l \in \llbracket 1, L \rrbracket} \frac{1}{h_{l} w_{l} c_{l}}\frob{\hat{F}^{(l)} - F^{(l)}}^2,\\
\mathcal{L}_{\text{style}}(\hat{\vx}, \vx_{\text{style}}) & = \sum_{l \in \llbracket 1, L \rrbracket} \frob{\hat{G}^{(l)} - G_{\text{style}}^{(l)}}^2,
\end{align}
where $\frob{\cdot}$ is the Frobenius norm, $\hat{G}^{(l)}{=}\hat{F}^{(l)\top}\hat{F}^{(l)} / (h_{l}w_{l}c_{l})$ and $G_{\text{style}}^{(l)}{=}F_{\text{style}}^{(l)\top}F_{\text{style}}^{(l)} / (\nh_{l} \nw_{l} c_{l})$.\footnote{For simplicity we formulate the losses across all layers of the network. In practice the losses are applied only to different subsets of layers that depend on the chosen architecture.}

The \emph{content reconstruction loss} $\mathcal{L}_{\text{content}}$ preserves the content from the original image during the style-transfer process, by keeping the features of the output image $\hat{\vx}$ close to the original ones of $\vx$. The \emph{style reconstruction loss} $\mathcal{L}_{\text{style}}$ is the squared Frobenius norm of the difference between the layer-wise Gram matrices of the output and style images and aligns the style statistics of the two images across layers.  

Beyond artistic style transfer, the ability to dissociate content from style in an image is potentially useful for other applications for visual perception.  For instance we could enable a model to recognize an object across different weather conditions by training it to ignore weather appearance, i.e., the style, and to focus on the useful features of the object, i.e., the content. We show the interest of such an approach in Figure~\ref{StyleTransfer}. Given a source image we use multiple realistic perturbations, e.g., fog, rain drops, sun glare, as style images to transfer to. We can see that each stylized image displays a particular visual appearance, which is also quantified with a high style reconstruction loss $\mathcal{L}_{\text{style}}$ between each pair of images. Interestingly, the content reconstruction loss $\mathcal{L}_{\text{content}}$ between them remains close to 0. 
This result motivates us to find a strategy to learn to manipulate the style information without damaging the content information and reach visual representations with increased robustness.

\subsection{Perturbation of style information}
\label{sec:removing-filters}
Consider an object of interest that we want to detect, e.g., a car. During the life-span of the vehicle, at the sensor level, the visual appearance of this type of object may vary significantly due to multiple factors: lighting, weather conditions, sensor aging and degradation. However, the considered object does not actually change dramatically during this time and we would expect the model to be robust to such variations. This means that the model must learn a robust representation of this object, resilient to multiple types of noise. 

In the following, we advance a few methods inspired by neural style transfer, to ensure robust representations by reducing style information from the feature maps of $f_{\vtheta}$. We will first test the feasibility of such an approach through three simple techniques. Note that in contrast to
typical style transfer, we do not require extra style images. From here onward we use only the information from the input image $\vx$.

\parag{$\text{Gram}_{\text{remove}}$.}
Here we neutralize the activations of the feature maps carrying most of the style information as following:
\begin{equation}
F^{(l)}_{i \in \phi^{(l)}} \leftarrow \vzero,
\end{equation}
where $\phi^{(l)}$ is the set of index filters corresponding to the top-$P$ values of the Gram matrix $G^{(l)}$. As an example, for $P{=}10\%$, the activation of the feature maps corresponding to the first 10\% highest values of the Gram matrix is set to $0$.

\parag{$\text{Gram}_{\text{weighting}}$.} 
Here we leverage the diagonal of the normalized Gram matrix to weight the activation maps:
\begin{equation}
F^{(l)} \leftarrow F^{(l)} [ \mathds{1} - \diag (G^{(l)}) ].
\end{equation}
The higher the values on the diagonal,
the bigger the attenuation. 
Conversely, feature maps with reduced style information undergo minimal changes.

\parag{$\text{Gram}_{\text{noise}}$.} 
In this case we perturb the same feature maps from $\phi^{(l)}$, as follows:
\begin{equation}
F^{(l)}_{i \in \phi^{(l)}} \leftarrow F^{(l)}_{i \in \phi^{(l)}} + \vepsilon^{(l)}G^{(l)} , \text{        } \vepsilon^{(l)} \sim \mathcal{N}(\vzero, \frac{1}{\tau}(\vsigma^{(l)})^2  ),
\end{equation}
where the standard deviation $\vsigma^{(l)}$ computed from $F^{(l)}$ and the noise weighting parameter $\tau$ are used for parameterizing a Normal distribution from which we sample the noise $\vepsilon^{(l)}$. We use $\vepsilon^{(l)}$ to perturb the style information in $G^{(l)}$ that will be further used to alter the feature maps $F^{(l)}$: the more style information carried, the more significant the perturbation.

\begin{figure}
\renewcommand{\captionfont}{\small}
\centering
\tiny
  \begin{tabular}{cc}
    Baseline & $\text{Gram}_{\text{remove}}$\\
    \includegraphics[width=3.5cm]{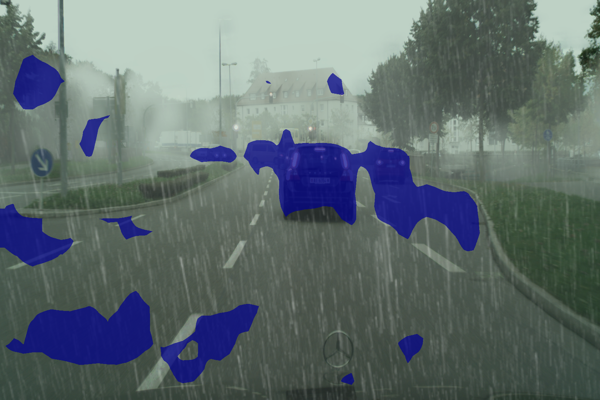} &
    \includegraphics[width=3.5cm]{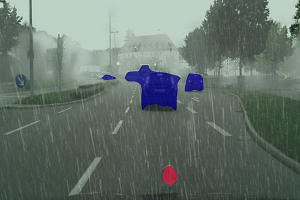}\\
    $\text{Gram}_{\text{noise}}$ & $\text{Gram}_{\text{weighting}}$ \\
    \includegraphics[width=3.5cm]{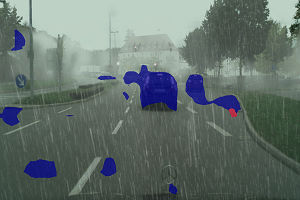}&
    \includegraphics[width=3.5cm]{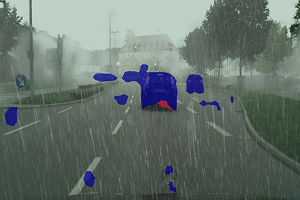}\\
  \end{tabular}
  \caption{\textbf{Qualitative results for style-perturbation filters 
  in semantic segmentation.} All models are trained in similar settings and here evaluated on an unseen rainy image. Vehicle predictions are depicted in blue color, and pedestrians in red.
 } 
 \label{fig:eval-filters}
\vspace{-10pt} 
\end{figure}

We validate this approach for the semantic segmentation task over images with unseen weather conditions, e.g., rain, and illustrate a qualitative example in Figure~\ref{fig:eval-filters}. We train all variants over the same data and training settings; we provide a detailed description in \S\ref{semseg}. From a qualitative evaluation, we can see that these filters outperform the baseline method in adverse weather conditions, e.g., rain with reduced light and visibility. These fairly simple filters confirm that discarding the style information can lead to performance improvements in unknown external conditions. This motivates us to design more advanced approaches to this effect. A drawback of the proposed filters is related to the per-feature map processing, which can zero-out an entire feature map without considering spatial information. 
Contrarily to style transfer where a global image style is sought, in our case the spatial information is actually key. For instance in foggy images, the visibility of the pixels will not be altered equally as regions closer to the camera are clearer, while further ones are partially or fully hidden by fog. Next, we attempt to design a trainable filter that would learn to diminish localized style information while preserving useful content.

\subsection{The \method~layer}
The previous style-perturbation strategies, albeit straightforward, are rather heuristic and could remove useful spatial and content information entangled with the style information. We seek a strategy that can be integrated during the training of the network, allowing it to better fit the training data and to learn to remove the style from input images without needing to compute Gram matrices at test time.
Geirhos et al.~\cite{geirhos2018imagenettrained} have shown that ImageNet-trained \dnns~do not leverage shape cues for recognizing objects, but rather over-rely on texture patterns. They mitigate this problem by using an aggressive data augmentation strategy, consisting in applying style transfer on training images using various types of popular style patterns. We could consider this strategy as well towards improving the robustness of our visual representations. However, we would first need to identify most of the typical style images to mirror the various weather conditions and sensor degradation. In addition, for each type of sensor, e.g., pinhole camera, fisheye camera, radar, this strategy must be revised and new suitable style images must be identified. We depart from this approach and propose a method for reducing
the style from the network features during training without using style-augmented training images. We propose a new layer, termed \emph{\method} that aims to attenuate or completely suppress style information from activations. Its filters are jointly trained with the main task, e.g., semantic segmentation, using an additional dedicated loss function. We describe below the structure of the {\method} layer and its corresponding loss.

In Figure~\ref{architecture} we illustrate the composition of a \method~layer that can be applied to any feature map as it acts as a residual connection. The input feature map goes through a sequence of Conv+ReLU layers that preserve its spatial resolution: $1\times1$ convolution layers at the entry and exit from this residual block (to compress and decompress channel-wise the feature maps and reduce number of parameters), and a middle $3\times3$ convolutional layer that aims to learn connections between filters (to some extent the style defined by the Gram matrix highlights information between filters).

We define the \emph{Gram loss} for learning the parameters $\vpsi$ of the \method~layers along with the rest of the parameters $\vtheta$:
\begin{equation}
\mathcal{L}_{\text{Gram}}(\vtheta,\vpsi) = 1-\frac{1}{L} \sum_{l \in \llbracket 1, L \rrbracket} \frac{\mean (G^{(l)}_{\text{in}} - G^{(l)}_{\text{out}})}{\max( G^{(l)}_{\text{in}})},
\end{equation}
where $G^{(l)}_{\text{in}}$, the normalized Gram matrix, is computed from the input feature maps of the $l^{\text{th}}$ \method~layer and $G^{(l)}_{\text{out}}$ from the output feature maps respectively.
Intuitively $\mathcal{L}_{\text{Gram}}$ encourages \method{} filters to remove useless information for the main task the network has been trained for. If there is no change in the feature maps the loss will top at $1$ as $G^{(l)}_{\text{in}}$ would be equal to $G^{(l)}_{\text{out}}$, while it will drop to $0$ if all information is removed, an edge case which is circumvented thanks to the joint training with the task loss. Note that a low $\mathcal{L}_{\text{Gram}}$ means most of the style information measured by the Gram matrix has been removed, but the content in the feature map is still there. In Figure~\ref{architecture} we show the effect of the \method{} filter on the activation maps of an image containing rain. In this case, we observe that many features belonging to the rain, which do not play a major role in the prediction, have been zeroed by the \method~filter.

\subsection{Training} \label{training_protocol}
The purpose of \method~layers is to reduce the style information embedded in the feature maps, thus enabling the network to learn other representations.
In our implementation, we first train a network for the main task to integrate both style and content information in the learned representation. 
Then, in a second stage, we activate the \method~layers and fine-tune the network on the same task and simultaneously attenuate the learned style information\footnote{Training from scratch with both losses, with conflicting impact on the style information, is unstable as the network needs both style and content to learn from in a first stage. \method~layers assume there is existing style information to reduce.
}. For this stage, the total loss reads:
\begin{equation}
\mathcal{L}_{\text{total}}(\vtheta, \vpsi) = \mathcal{L}_{\text{task}}(\vtheta, \vpsi) + \alpha \mathcal{L}_{\text{Gram}}(\vtheta, \vpsi),
\end{equation}
where $\alpha$ is a hyper-parameter for balancing the two losses. 
In our experiments we use $\alpha{=}0.1$.

To summarize, the training protocol consists in the following steps:
\begin{enumerate}
\item Train the network $f_{\vtheta}(\cdot)$ on its own task with $\mathcal{L}_{\text{task}}(\vtheta)$; 
\item Add \method~layers with parameters $\vpsi$ at various levels in the network. For ResNet backbones~\cite{he2016deep}, we connect \method~layers after each residual block\footnote{In the presence of \method~layers $\vpsi$, we denote the network by $f_{\vtheta, \vpsi}(\cdot)$};
\item Fine-tune all layers of $f_{\vtheta, \vpsi}(\cdot)$ with $\mathcal{L}_{\text{total}}(\vtheta, \vpsi)$.
\end{enumerate}

\begin{table}[t!]
\centering
\renewcommand{\captionfont}{\footnotesize} 
\caption{\textbf{Semantic segmentation (\emph{road}, \emph{vulnerables}, \emph{vehicles}) across weather conditions (mIoU).} All models are trained on Cityscapes~\cite{Cityscapes} 
and evaluated on Cityscapes, Foggy Cityscapes~\cite{dai2019curriculum}, and Rain Cityscapes~\cite{RainyCityscapes}. $^{\dag}$: our implementation.
}
\resizebox{\columnwidth}{!}{%
\begin{tabular}{l | c c c| c c c |c c c}
\toprule
       &\multicolumn{3}{c|}{\textbf{Cityscapes}} & \multicolumn{3}{c|}{\textbf{Foggy Cityscapes}} &
      \multicolumn{3}{c}{\textbf{Rain Cityscapes}}\\
\textbf{Model} & road & vul. & veh. &  road & vul. & veh. &  road & vul. & veh.\\
\midrule
\multicolumn{10}{l}{\textbf{Baseline}}\\
\;per-class   & 92.2 & 67.3 & 86.8 & 88.0 & 58.2 & 76.6 & 31.8 & 59.3 & 64.0\\
\;\;\;\;overall   & \multicolumn{3}{c|}{84.8} & \multicolumn{3}{c|}{78.0} & \multicolumn{3}{c}{55.2}\\
\midrule      
\multicolumn{10}{l}{\textbf{PbN~\cite{kamann2020increasing}$^{\dag}$}}\\
\;per-class  & 92.9 & 70.6 & 87.8 & \textbf{92.2} & 67.1 & 82.7 &
47.2 & 60.6 & 67.3\\
\;\;\;\;overall   & \multicolumn{3}{c|}{86.2} & \multicolumn{3}{c|}{83.5} & \multicolumn{3}{c}{61.6}\\
\midrule 
\multicolumn{10}{l}{\textbf{StyleLess}}\\
\;per-class & \textbf{93.6} & \textbf{72.6} & \textbf{89.4} & 92.1 & \textbf{67.2} & \textbf{84.0} & \textbf{63.0} & \textbf{64.5} & \textbf{74.3}\\
\;\;\;\;overall   & \multicolumn{3}{c|}{\textbf{87.4}} & \multicolumn{3}{c|}{\textbf{84.0}} & \multicolumn{3}{c}{\textbf{69.9}}\\
\bottomrule
\end{tabular}%
}
\label{SemanticSegResutsFoggy}
\vspace{-8pt}
\end{table}

\begin{table}[t]
\centering
\renewcommand{\captionfont}{\footnotesize} 
\caption{\textbf{Object detection  across weather conditions (mAP).} Models are trained for eight-class detection on Cityscapes~\cite{Cityscapes} and evaluated on Cityscapes and Foggy Cityscapes~\cite{dai2019curriculum}. $^{\ddag}$: trains with unlabeled foggy images.}
\resizebox{\columnwidth}{!}{%
\begin{tabular}{l|cccccccc|c}
\toprule
\textbf{Model} & \rot{bike} & \rot{bus} & \rot{car} & \rot{motor} & \rot{person} & \rot{rider} & \rot{train} & \rot{truck} & \textbf{mAP}\\
\midrule
SSD-300~\cite{SSD} & 29.2 & 22.0 & 50.9 & 14.6 & 30.2 & 29.8 & 9.1 & 11.9 & 24.7\\
\midrule
Advent~\cite{advent}$^{\ddag}$ & 23.1 & \textbf{37.1} & 39.6 & \textbf{21.3} & 17.6 & 25.0 & \textbf{25.9} & \textbf{20.0} & \textbf{26.2}\\
\midrule
\method & \textbf{31.3} & 25.2 & \textbf{51.6} & 16.1 & \textbf{32.8} & \textbf{30.9} & 9.1& 16.2 & \textbf{26.6}
\\
\bottomrule
\end{tabular}%
}
\label{ODResluts}
\end{table}

%% file: 4-experiments.tex
\section{Experiments}

We evaluate our method on two key tasks for autonomous driving, \emph{semantic segmentation} and \emph{object detection}, over several instances of the Cityscapes dataset~\cite{Cityscapes} with various challenging conditions: clean, foggy~\cite{dai2019curriculum}, rainy~\cite{RainyCityscapes}, corrupted.

\subsection{Datasets}
Cityscapes~\cite{Cityscapes} consists of $3,475$ images ($2,975$ training, $500$ validation) with pixel-level annotations, from which we can extract bounding box annotations. We further consider additional variants of this dataset: Foggy Cityscapes~\cite{dai2019curriculum} with realistic depth-aware simulated fog images with different densities ($1,500$ validation images derived from $500$ images with $3$ fog densities), and Rain Cityscapes~\cite{RainyCityscapes} with realistic rain effects ($1,188$ validation images obtained from $33$ images, each with $3$ different densities and $12$ different rain patterns). Furthermore, similarly to \cite{kamann2020increasing}, we generate a corrupted variant of Cityscapes, called Cityscapes-C, using $15$ image corruption strategies from \cite{hendrycks2019benchmarking}. Cityscapes-C amounts to $7,500$ images.

\begin{figure}[t!]
\renewcommand{\captionfont}{\small} 
\begin{center}
\begin{tabular}{ c c }
    \includegraphics[width=3.8cm]{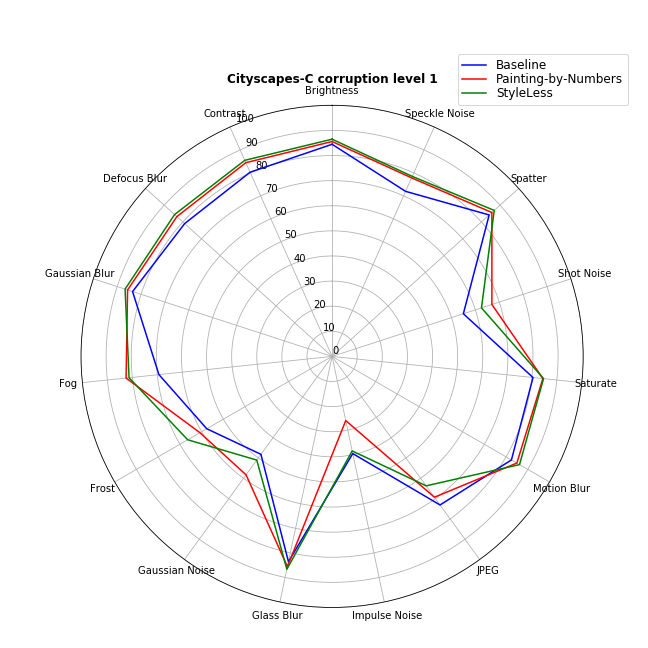} & 
    \includegraphics[width=3.8cm]{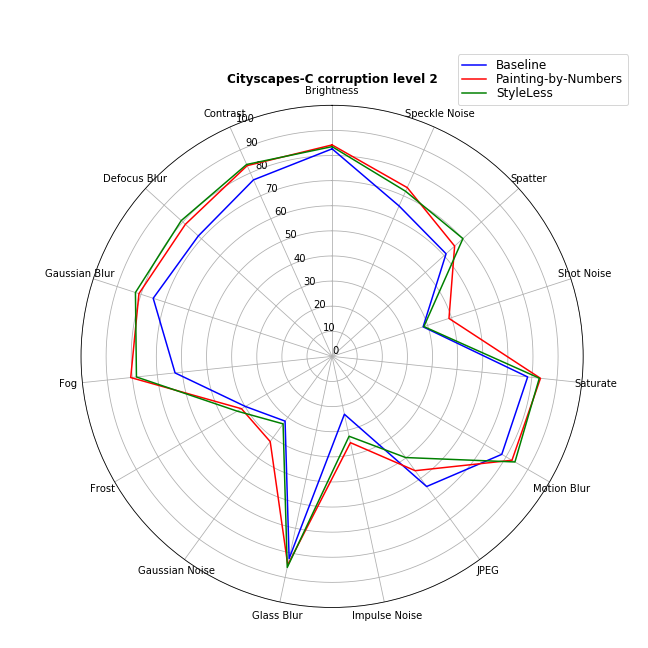}\\
    \includegraphics[width=3.8cm]{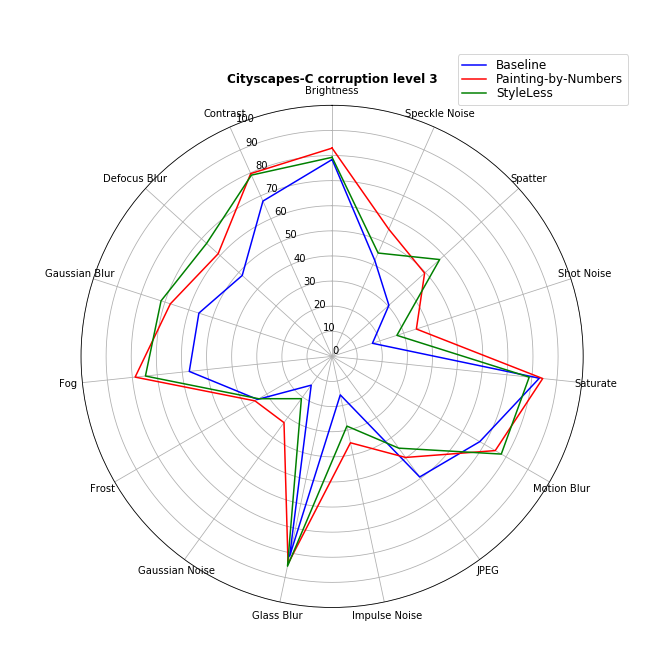} & 
    \includegraphics[width=3.8cm]{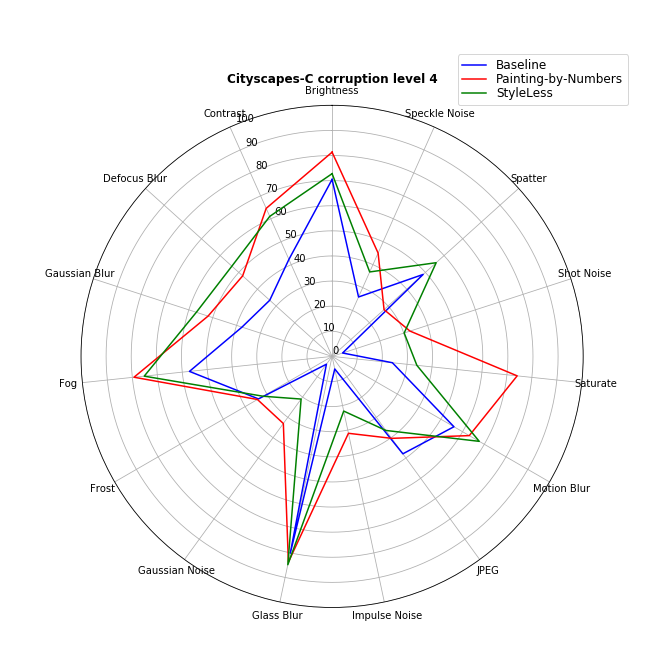}\\
\end{tabular}
\end{center}
\caption{\textbf{Semantic segmentation (\emph{road}, \emph{vulnerables}, \emph{vehicles}) across image corruptions (mIoU).} Models are trained on Cityscapes~\cite{Cityscapes} and evaluated on Cityscapes-C. Results are over corruption levels 1-4.}
\label{spyder_chart}
\vspace{-5mm}
\end{figure}

\begin{figure*}[t!]
\renewcommand{\captionfont}{\small}
\centering
\scriptsize
  \begin{tabular}{c c c c c}
    \toprule
    & Spatter Noise & Glass Blur& Frost & Gaussian Blur\\
    \raisebox{2.0\normalbaselineskip}[0pt][0pt]{\rotatebox[origin=c]{90}{GT}} &
    \includegraphics[width=3cm]{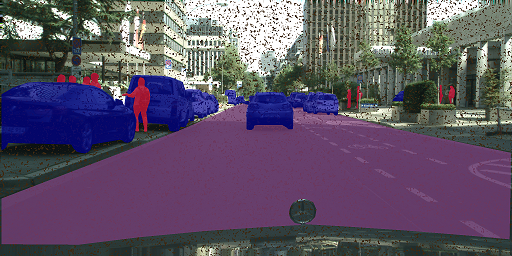} & 
    \includegraphics[width=3cm]{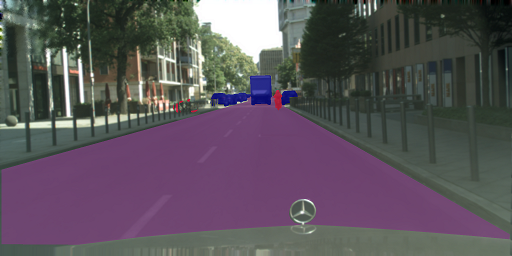} &
    \includegraphics[width=3cm]{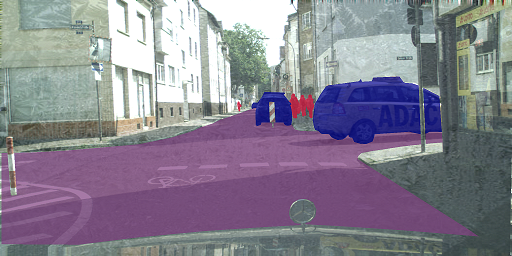} & 
    \includegraphics[width=3cm]{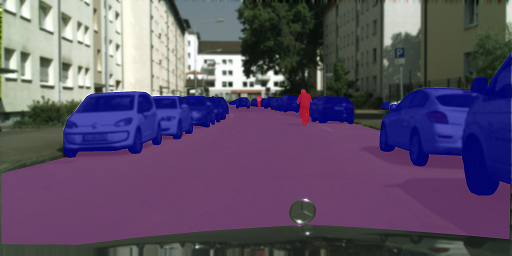} \\
    
    \raisebox{2.5\normalbaselineskip}[0pt][0pt]{\rotatebox[origin=c]{90}{Baseline}}
    &
    \includegraphics[width=3cm]{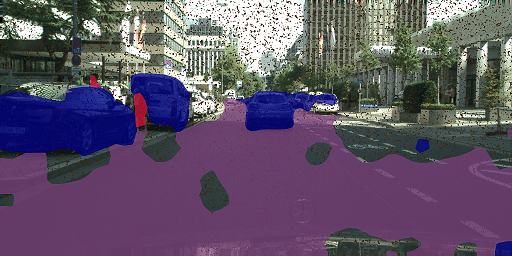} & 
    \includegraphics[width=3cm]{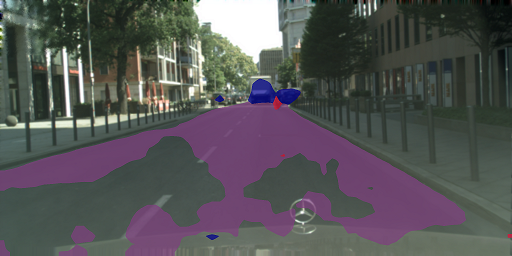} &
    \includegraphics[width=3cm]{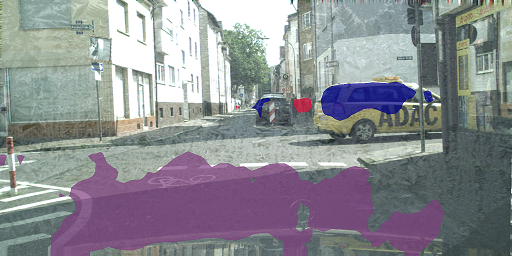} & 
    \includegraphics[width=3cm]{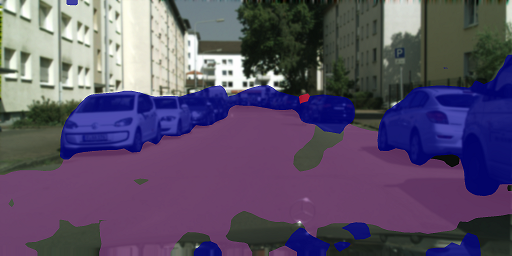} \\
    
    \raisebox{2.5\normalbaselineskip}[0pt][0pt]{\rotatebox[origin=c]{90}{PbN}}
    &
    \includegraphics[width=3cm]{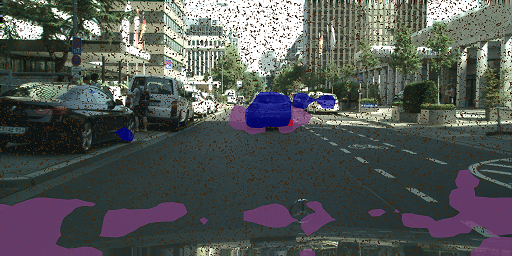} & 
    \includegraphics[width=3cm]{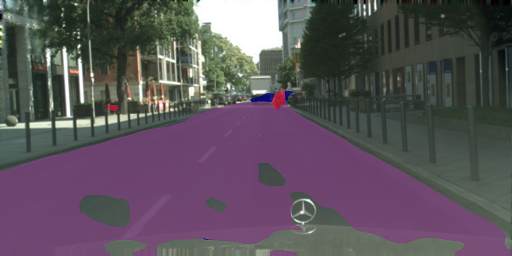} &
    \includegraphics[width=3cm]{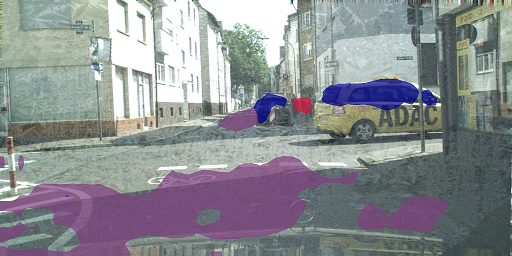} & 
    \includegraphics[width=3cm]{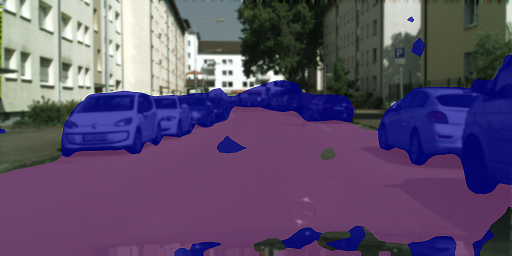} \\
    
    \raisebox{2.3\normalbaselineskip}[0pt][0pt]{\rotatebox[origin=c]{90}{StyleLess}}
    &
    \includegraphics[width=3cm]{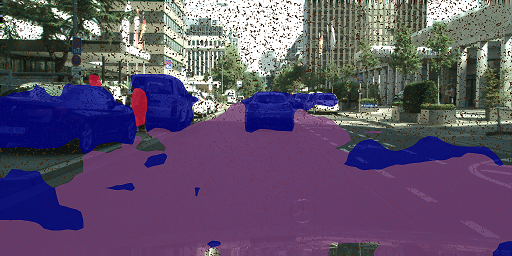} & 
    \includegraphics[width=3cm]{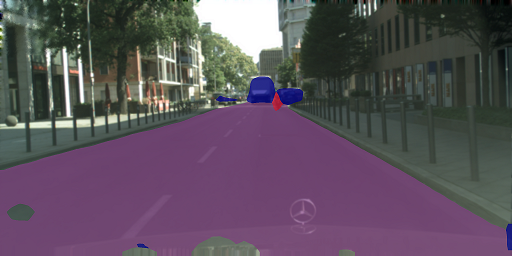} &
    \includegraphics[width=3cm]{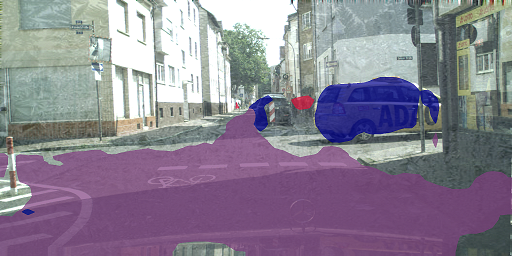} & 
    \includegraphics[width=3cm]{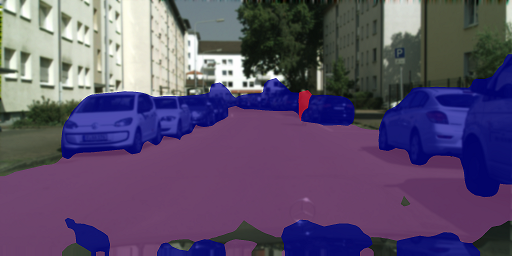} \\
    
    &
    \includegraphics[width=3cm]{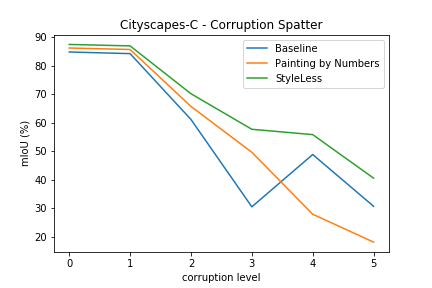} & 
    \includegraphics[width=3cm]{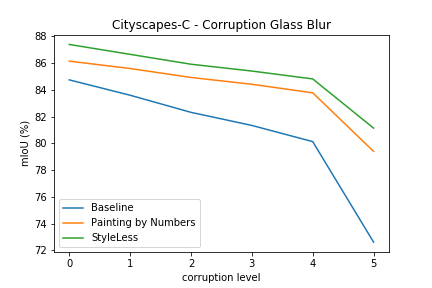} &
    \includegraphics[width=3cm]{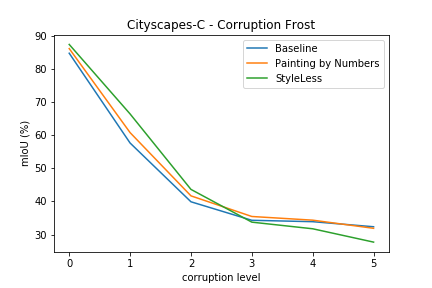} & 
    \includegraphics[width=3cm]{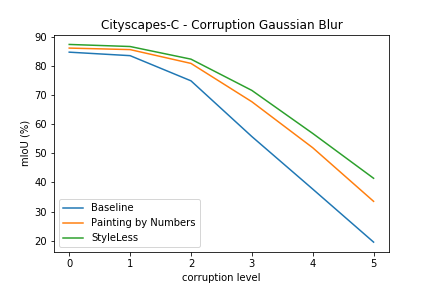} \\
    
  \end{tabular}
  \caption{
  \textbf{Qualitative results for semantic segmentation on Cityscapes-C for \emph{Spatter Noise}, \emph{Glass Blur}, \emph{Frost}, and \emph{Gaussian Blur} corruptions.} ($1^{st}$ row) Ground Truth; ($2^{nd}$ row) Baseline; ($3^{rd}$ row) Painting-by-Numbers~\cite{kamann2020increasing}; ($4^{th}$ row) \method; ($5^{th}$ row) aggregated mIoU scores across corruption levels.
  }
\label{fig:qualitative}
\vspace{-10pt}
\end{figure*}

\subsection{Semantic Segmentation} \label{semseg}

\noindent \textbf{Implementation details.} We consider the popular DeepLabV3 architecture~\cite{DeepLabV3} for its excellent trade-off between run-time and predictive performance. We combine it with a ResNet18 backbone~\cite{he2016deep} pretrained on ImageNet~\cite{russakovsky2015imagenet}, which is more compact and closer to typical \dnns~running on vehicle hardware. We follow the steps describe in \S\ref{training_protocol}. We first train the network for the main task, semantic segmentation on Cityscapes training set, with the following settings: random crops of $512\times512$, SGD with polynomial learning rate annealing with initial learning rate $0.01$, momentum of $0.9$ and weight decay of $10^{-4}$. We then activate the \method{} layers (about $2\%$ of the network parameters) and fine-tune the entire network with $\mathcal{L}_{\text{task}}$ and $\mathcal{L}_{\text{Gram}}$ over the same Cityscapes set. The learning rate for the newly added \method{} layers is $10$ times higher than for the backbone. In order to better highlight the impact of the \method{} layer, we regroup the classes into \emph{road}, \emph{vehicle} (car, truck, bus, trailer) and \emph{vulnerable} (person, rider), leaving the remaining classes as background. 

\noindent \textbf{Baselines.} We validate our method against the simple baseline method trained on Cityscapes and against the recent Painting-by-Numbers (PbN)~\cite{kamann2020increasing} approach. We implemented it ourselves following the description from the paper.\footnote{Kamann et al.~\cite{kamann2020increasing} use DeepLabV3+ with ResNet50 backbone and train on 19 Cityscapes classes. Here, in order to comply with typical automotive constraints, we use DeepLabV3 with ResNet18 backbone for 4 groups of classes (background, road, vehicles, vulnerables).}

\noindent \textbf{Results.} We evaluate all methods and report performance in terms of mean intersection-over-union (mIoU) scores ($\%$). We first evaluate our simple handcrafted filters for removing style (\S\ref{sec:removing-filters}) and as expected they display small gains ($+1\%$) in performance w.r.t. baseline, emphasizing the need of a more elaborated approach. We report our results in unseen weather conditions, fog and rain, in Table~\ref{SemanticSegResutsFoggy}. We can see that \method{} consistently improves over the baseline with a comfortable margin and most of the times over PbN. In Figure~\ref{spyder_chart}, we illustrate the results on Cityscapes-C with unseen synthetic image corruptions. Here, the gap between \method{} and PbN is smaller, with the two methods reaching similar performances. We emphasize that \method{} does not rely on specific image annotations, e.g., segmentation map, as PbN, and can be directly applied to various tasks and sensor modalities. We present a few qualitative results from Cityscapes-C in Figure~\ref{fig:qualitative}. We can observe that \method{} outperforms the baseline in all cases, and has an edge over PbN on \emph{spatter noise} and \emph{glass blur} across corruption levels.

\begin{figure}[t!]
    \renewcommand{\captionfont}{\small}
    \centering
    \includegraphics[width=8cm]{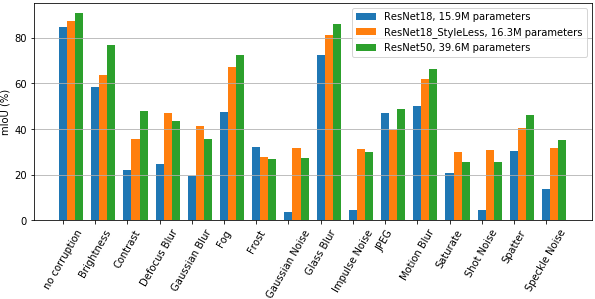}
    \caption{\textbf{Semantic segmentation (\emph{road}, \emph{vulnerables}, \emph{vehicles}) performance}. Comparison of different backbones on clean and corrupted images. Image corruption level is 5.
    }
    \label{fig:ablation-c}
\vspace{-8pt}
\end{figure}

\noindent \textbf{Discussion.} Inserting \method{} layers in a backbone increases slightly the capacity of the network with the newly added parameters ($+2\%$ more parameters for ResNet18). We assess whether the performance boost of \method{} is due to the higher capacity or to the style attenuation effect. To this end, we consider backbones of varying sizes (ResNet18, ResNet18+\method{} layers, ResNet50), train them on Cityscapes and evaluate on Cityscapes and Cityscapes-C. We report results in Figure~\ref{fig:ablation-c}. In spite of having two-times fewer parameters, ResNet18+\method\  closely follows ResNet50, in particular on the challenging Cityscapes-C, and outperforms by a larger margin ResNet18 with similar capacity. This confirms the effectiveness of our approach for improving robustness with small capacity overhead.

\subsection{Object Detection}
We illustrate the versatility of \method{} across tasks by validating it on object detection.

\noindent \textbf{Implementation details.} We use a single stage detector, SSD~\cite{SSD}, with a pre-trained VGG-16 backbone~\cite{VGG} . We train on eight-class detection on Cityscapes with the following settings: $300 \times 300$ images, SGD with learning rate $10^{-3}$ and multi-step scheduler with $\gamma=0.1$ and milestones [$120$, $160$], momentum $0.9$,  weight decay $5\times10^{-4}$.

\noindent \textbf{Baselines.} We compare against the simple baseline and against Advent~\cite{advent}, an unsupervised domain adaptation approach that can learn from unlabeled target-domain images.

\noindent \textbf{Results.} We evaluate on unseen weather conditions from Foggy Cityscapes and report mean average-precision (mAP) scores across classes in Table~\ref{ODResluts}. \method{} outperforms the baseline on all classes and does slightly better than Advent, in spite of the advantage of Advent that ``sees'' unlabeled foggy images during training. The two methods improve on different object classes, hinting that they can potentially complement each other. 

%% file: 5-conclusion.tex
\section{Conclusion}

We introduced a new \method~layer that enables \dnns~to learn robust visual representations. This approach offers substantial gains on various corruptions, including adverse weather conditions.
\method~does not rely on specific data augmentations, thus no prior sensor knowledge, in order to separate style information from content information. In future work, we plan to evaluate the efficiency of our \method~layer on other sensor modalities such as radar or lidar where labelled data is costly. We expect a significant gain towards improving robustness in unseen situations and towards reducing annotation and hardware costs.